\documentclass[]{spie}  

\newcommand{\B}{\fontseries{b}\selectfont} 

\usepackage{amsmath,amsfonts,amssymb}
\usepackage{graphicx}
\usepackage[colorlinks=true, allcolors=blue]{hyperref}
\usepackage{textcmds} 

\title{Data augmentation and pre-trained networks for extremely low data regimes unsupervised visual inspection}

\author[1]{Pierre Gutierrez}
\author[1]{Antoine Cordier}
\author[2]{Thaïs Caldeira}
\author[1]{Théophile Sautory}

\affil[1]{Scortex, 22 rue Berbier du Mets, Paris, France}
\affil[2]{School of Electrical and Computer Engineering, University of Campinas, Brazil}

\authorinfo{For further author information, send correspondence to P. Gutierrez\\E-mail: pgutierrez@scortex.io}

\begin{document} 
\maketitle

\begin{abstract}
The use of deep features coming from pre-trained neural networks for unsupervised anomaly detection purposes has recently gathered momentum in the computer vision field. In particular, industrial inspection applications can take advantage of such features, as demonstrated by the multiple successes of related methods on the MVTec Anomaly Detection (MVTec AD) dataset. These methods make use of neural networks pre-trained on auxiliary classification tasks such as ImageNet. However, to our knowledge, no comparative study of robustness to the low data regimes between these approaches has been conducted yet. For quality inspection applications, the handling of limited sample sizes may be crucial as large quantities of images are not available for small series. In this work, we aim to compare three approaches based on deep pre-trained features when varying the quantity of available data in MVTec AD: KNN, Mahalanobis, and PaDiM. We show that although these methods are mostly robust to small sample sizes, they still can benefit greatly from using data augmentation in the original image space, which allows to deal with very small production runs.
\end{abstract}


\section{Introduction}
\label{sec:intro}  

Deep learning is currently revolutionizing automated visual inspection. Practitioners usually adopt either a supervised learning approach (where the system is trained with labelled defective images) or an unsupervised learning one (where any anomaly on a test image is considered as a potential defect). In the case of very limited series and custom-manufactured goods, the amount of available data to train an automatic inspection system is often quite low (typically, below 50 images). This prevents the adoption of a supervised approach, since defective images cannot be collected in sufficient quantities for training. At the same time, training the system should not impact the production cycle times. For small series, this means that only a few minutes, or even less, are available for training.

A recent unsupervised approach \cite{rippel2020modeling} from the literature leverages deep features from pre-trained networks for anomaly detection. Using such a network as a frozen feature extractor can effectively reduce training time, as well as the required amount of training data. In this work, we focus on the extremely low data regime. First, we review and benchmark several unsupervised methods from the literature which make use of these pre-trained networks, by studying the robustness of these methods to the low data regimes. To improve this robustness, we then propose applying data augmentation, by synthetically generating modified images: we find that data augmentation has a strong and positive impact in such regimes. 

Results are reported on the publicly available MVTec AD dataset \cite{bergmann2019mvtec}. An average AUC of above 0.90 can be achieved with only a few images, which allows to tackle very small series. One may notice that the AUCs in low-data regimes for the reported methods are on par with the ones obtained with autoencoders in maximum-data regimes.\cite{bergmann2019mvtec}

\section{Related work}
\label{sec:related_work}  

Quality control can be automated using supervised deep learning \cite{ren2017generic, cordier2021active}. To do so, \textbf{convolutional neural networks (CNNs)} such as U-Net \cite{ronneberger2015u} or RetinaNet\cite{lin2017focal} are trained to explicitly recognize known defects. One of the major downsides of this method is that it may require a high volume of annotated images. Another drawback is that it cannot handle unusual or theoretical defects. Two main approaches are being proposed to solve these issues: simulations \cite{zambal2019end, gutierrez2021synthetic} and anomaly detection (via unsupervised or semi-supervised learning). In this work, we focus on the latter.

\textbf{Anomaly detection} in general and on images in particular has been a very popular field in the past few years. We thus focus this review on the papers that are the closest to our task at hand, and refer the reader to a broader review of literature \cite{pang2020deep} for further details. Generally speaking, most approaches fall in any of the three following categories: reconstruction of the input, classification of anomalies and modelling of normality.

Autoencoders for \textbf{image reconstruction} have been among the first methods to make use of deep learning for anomaly detection \cite{bergmann2019mvtec,huang2019attribute}. The idea is to reconstruct the input image using a network that was trained on normal data solely, before computing the anomaly score as a distance between the input and the output reconstruction. Because the simple pixel-wise L2 difference is a poor choice of similarity measure between images, the approach has been extended using structural similarity\cite{bergmann2018improving}, perceptual loss, variational autoencoders (VAEs) \cite{baur2021autoencoders} and generative adversarial networks (GANs) \cite{schlegl2017unsupervised}. For an extensive overview and experimental comparison of some of these methods, see Baur and al. \cite{baur2021autoencoders}.

Another possibility is to \textbf{leverage the classification confidences of a CNN} trained in a supervised or self-supervised fashion in order to predict anomalies. For example, data augmentation techniques can be used to assess a network confidence zone: in GeoTrans \cite{golan2018deep}, the authors propose to learn and predict geometric transformations performed on an input image, like translations or rotations. The intuition is that at test time, if an image is out-of-distribution, the network should not be able to predict the transformation correctly. This is in line with the idea that self-supervised techniques can improve uncertainty estimation, even as an auxiliary task\cite{hendrycks2019using}. The idea from GeoTrans has been extended in GOAD \cite{bergman2020classification}, using ideas from the deep one-class \cite{perera2019learning} literature. However, the performance of both of these approaches depends on what can be considered as acceptable data augmentations with regards to the dataset of interest. CSI \cite{tack2020csi} builds on this idea and improves the method by using contrastive learning and defining what are acceptable and out-of-distributions transformations. To some extent, this can be seen as a way to do outlier exposure \cite{hendrycks2018deep}, i.e. a supervised training of the network so that it learns to separate inliers from outliers (which can come from either synthetic generation or external datasets).

In order to \textbf{model normality}, and inspired by the one-class SVM technique \cite{scholkopf1999support}, Deep SVDD \cite{perera2019learning} projects the input data on an hyper-sphere. The main issue with the Deep SVDD method is that the network does not necessarily learn discriminative features. One way to fix this is to extend Deep SVDD to the semi-supervised case (Deep SAD) \cite{ruff2019deep,ruff2020rethinking}, potentially using synthetic anomalies \cite{liznerski2020explainable}. Another way to force the network to be descriptive is to simultaneously learn an auxiliary supervised task on another unrelated dataset \cite{perera2019learning}. 

A recent trend in modelling normality for anomaly detection is to directly take advantage of \textbf{deep feature representations from a pre-trained network}. Using a pre-trained CNN, images are encoded in the embedding space of the network where traditional anomaly detectors can then be used. This guarantees descriptiveness of the features, and avoids mode collapses since the pre-training is performed on a large, generalist dataset such as ImageNet\cite{5206848}. Typical methods for computing the final anomaly score in the embedding space are: distance to the k-nearest neighbors (KNN) \cite{bergman2020deep}, Mahalanobis distance \cite{rippel2020modeling}, and kernel density estimation \cite{erdil2020unsupervised}. Because these methods only provide an anomaly score at the image level and do not necessarily offer anomaly localization, they have been improved in several ways. SPADE\cite{cohen2020sub} extends the KNN approach by looking at the nearest neighbors locally, in the pixel-embedding space. Correspondingly, PaDiM \cite{defard2020padim} extends the Mahanalobis approach by locally modeling patch-wise gaussian covariances. Finally, in “Deep Feature Reconstruction” (DFR)\cite{yang2020dfr}, Yang et. al. use a reconstruction autoencoder in the pixel-embedding space of the pre-trained feature extractor. Note that the idea of feature reconstruction was first proposed without the need of a frozen pre-trained network. It was introduced in the “Uninformed Students” paper \cite{bergmann2020uninformed}, where an ensemble of CNNs is trained to mimic a teacher network. The idea has since been extended by using several feature maps from a single network \cite{wang2021student, salehi2020multiresolution}, as well as by blurring the input \cite{choi2019novelty}.

The methods benefiting from frozen pre-trained networks often perform very well because they cannot forget the richness of the pre-trained feature representations, which often happens when fine-tuning such networks on different data due to catastrophic forgetting \cite{goodfellow2013empirical}. By taking advantage of descriptive features, combined with the compactness that the traditional anomaly detectors bring, these methods allow to get state-of-the-art performance on many standard anomaly detection datasets for computer vision, such as MVTec. It is worth noting that techniques such as KNN, Mahalanobis, and PaDiM are also very fast to train and to test  (seconds or minutes) compared to other methods that require an additional CNN training, such as autoencoders or self-supervised networks (hours or days). This makes the said methods prime for small series in quality inspection use cases.

\section{Materials and methods}
\label{sec:methods}  

\subsection{Pre-trained networks for anomaly detection}
\label{subsec:pre-trained}
\begin{figure}
    \centering
    \includegraphics[height=8.5cm]{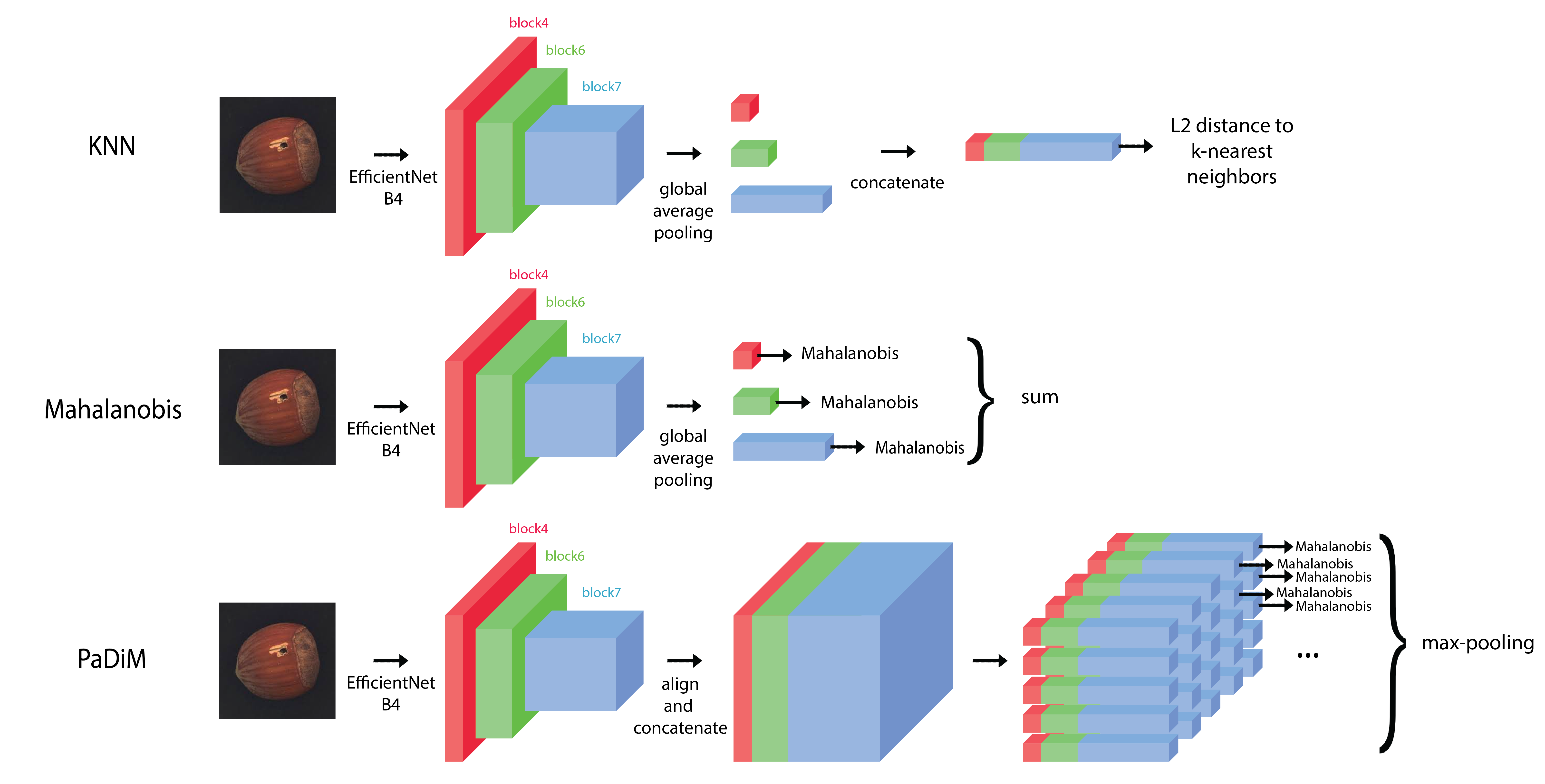}
    \caption{Schematic overview of the three pre-trained network based methods that we compare in this work: KNN, Mahalanobis, and PaDiM. In the KNN and PaDiM approaches, features from the different feature levels are concatenated. In both the KNN and the Mahalanobis approaches, global average pooling aggregates the information spatially and allows to manipulate smaller vectors. In PaDiM, spatial aggregation is done \textit{a posteriori}, by taking the maximum local Mahalanobis distance over all pixel-embeddings.}
    \label{fig:overview_three_methods}
\end{figure}

In this article, as we focus on very small training sample sizes, we only consider methods leveraging pre-trained feature extractors. We compare the following methods: KNN \cite{bergman2020deep} (and therefore SPADE\cite{cohen2020sub}, for which the image anomaly score is identical), Mahalanabis\cite{rippel2020modeling} and PaDim\cite{defard2020padim}. Figure \ref{fig:overview_three_methods} summarizes all three approaches. All methods output an anomaly score. In practice, a threshold is set to classify anomalous from normal images.

\subsubsection{KNN}
The DN2 (Deep Neareast Neighbor Anomaly Detection) method developed by Bergman et al.\cite{bergman2020deep} uses a k nearest-neighbor approach in the global-average pooled embedding space of a feature extractor for anomaly detection: a training set constituted of normal images is first embedded (and pooled), before embedding (and pooling) test images. The anomaly score associated with a given test image $y$ can then be calculated using the average euclidean distance of its associated embedded vector $f_{y}$ to the $k$ nearest training embeddings $N_{k}(f_{y})$:
\begin{equation}
d(y) = \frac{1}{k} \sum_{f\in N_{k}(f_{y})}{||f-f_{y}||^2}
\end{equation}
While in the original paper, only the last feature level embeddings are used, the work is extended with SPADE (Sub-Image Anomaly Detection with Deep Pyramid Correspondences)\cite{cohen2020sub}, where spatial features coming from multiple levels are aligned and concatenated for computing an additional pixel-wise score with a second KNN. The method will not be discussed in detail here since its resulting image score is in fact identical to the original KNN\cite{bergman2020deep} implementation. However, the idea of using feature levels coming from different scales is of interest, as it has been shown to bring value for anomaly detection using pre-trained networks.\cite{rippel2020modeling} Consequently, and for fairness of comparison with the two other methods (which also use multiple feature levels), for the KNN approach we decide to compute every $f$ embedding vector by concatenating the global average pooled features coming from the different levels we preliminarily chose, in order to obtain the final embedded vector. In this work, we set the number of nearest neighbors $k$ to 1, as it gives the best results in our experiments.

\subsubsection{Mahalanobis}
Instead of using euclidean distance to the nearest neighbors in the latent space, Rippel et al.\cite{rippel2020modeling} proposed to model the embedded training set distribution as a multivariate gaussian and to use the Mahalanobis distance as the anomaly score. After computing the embedding vectors for the training set using an EfficientNet feature extractor, a covariance matrix for the features is estimated. The Mahalanobis distance of an embedded test set sample $f_{y}$ to the mean of the embedded training set distribution $\mu$ can then be computed:
\begin{equation}
d(y) = \sqrt{(f_{y} - \mu)^{T} S^{-1} (f_{y} - \mu)}
\end{equation}
where $S$ is the estimated covariance matrix. In practice, in order to make use of the multiple available feature levels, a covariance matrix is estimated for each feature level that was preliminarily chosen, and the final anomaly score for a given image $y$ is calculated as the sum of the Mahalanobis distances for all feature levels. Alternatively (and similarly to what we do for the KNN approach), embeddings from the different feature levels can first be concatenated after pooling, and a single covariance matrix for all feature levels can then be estimated to compute a unique Mahalanobis distance as the anomaly score. In practice, we find that both methods lead to similar results (data not shown). Hence, we decide to stick with the original implementation, i.e. the sum of the different Mahalanobis distances over the multiple feature levels.

The advantage of using the Mahalanobis distance is that unlike classical euclidean distance, components with the highest variance will be weighting less than their lowest variance counterparts in the final computed distance.  In that regard, work from Kamoi et al.\cite{kamoi2020mahalanobis} suggests that the few features that explain most of the variance are not the ones that contribute to anomaly detection in high-dimensional cases, but rather the ones useful for the original classification pre-training task. This makes Mahalanobis distance a metric of choice when working with distributions in multi-dimensional spaces. However, the drawback is that it implicitly assumes the underlying embedded training sample distribution to be unimodal (i.e. distributed uniformly around a unique center), which might not be true. Note that the Mahalanobis distance is identical to the euclidean distance in the case where the co-variance matrix $S$ is equal to the identity.

To compute the covariance matrix, we use two different estimators: Empirical, and Ledoit-Wolf\cite{LEDOIT2004365}. While the Empirical (or maximum likelihood) estimation works well given enough samples are provided, Ledoit's method goes further by shrinking the empirical covariance matrix: diagonal coefficients of the covariance matrix are artificially boosted, lessening the impact of the non-diagonal ones. This is expected to be valuable in low data regimes, where the number of available datapoints can become lower than the number of features for which to compute the covariance matrix. Errors in estimations will therefore be reduced, and invertibility of the matrix is expected to improve.

\subsubsection{PaDiM}
PaDiM (Patch Distribution Modeling Framework for Anomaly Detection and Localization)\cite{defard2020padim} can be seen as a pixel-wise extension of the Mahalanobis approach, where one covariance matrix is estimated per spatial embedded location. Similarly to what is done for the SPADE\cite{cohen2020sub} anomaly localization module, the extracted spatial embeddings are first aligned and concatenated over feature levels. Because these concatenated spatial embeddings are not global-average-pooled like for the KNN or Mahalanobis method, one covariance matrix $S_{i,j}$ can then be estimated per spatial location (or patch) $(i, j)$ in the concatenated embedding space, leading to one anomaly score per location (or patch). The final image anomaly score can then be taken as the maximum of the anomaly scores for all locations:
\begin{equation}
d(y) = \max_{i,j}{\sqrt{(f_{y_{i,j}} - \mu_{i,j})^{T} S_{i,j}^{-1} (f_{y_{i,j}} - \mu_{i,j})}}
\end{equation}
The advantage of this approach is that it exploits direct correlations between the features at different scales through the concatenation of feature levels, and allows to localize anomalies on heatmaps without relying on gradient-based visual explanations such as Grad-CAM\cite{DBLP:journals/corr/SelvarajuDVCPB16}. PaDiM outperforms other existing methods on the MVTec dataset, but has a significant drawback: its number of covariance matrices (equal to the number of spatial locations in the aligned and concatenated embedding space). This can significantly increase both training and inference times in practice, due to the estimation of the covariance matrices and the computation of corresponding Mahalanobis distances, respectively.

\subsection{Data augmentation}
\label{subsec:data_augm}

Data augmentation is used in the original Mahalanobis method \cite{rippel2020modeling} paper. The authors find that although it does not improve the detection performances significantly, data augmentation still is paramount to automatically set a working point (threshold). Data augmentation has been widely used as a regularization process in the training of neural networks to improve generalization, but it is not obvious that it can help when the feature extractor is frozen. Notably, there is \textit{a priori} no reason that data augmentation maintains the embeddings gaussianity assumption of the Mahalanobis method. In this work, we use data augmentation as it is susceptible to help each of the three methods detailed above in the extremely low data regimes by enriching the training set with new samples.

Because images invariances are different for each class of object and texture of the MVTec dataset, we use different data augmentations for each. We design them manually, by visualizing augmented images from the training sets. The set of available transformations from which we pick are: vertical and horizontal flips, translations, rotations, zoom, and brightness change. For brightness, we either add to each image a random value for all pixels, or multiply each image by a random multiplicative factor. For rotations in texture classes, we mostly use multiples of 90° since they do not require any type of padding for squared images. For each transformation, we define whether they are applied or not, and if yes in which range. When selected for a given class, these data augmentations are applied systematically: for example, for the capsule class, each image is translated by a different randomly sampled number of pixels in the range [-10, 10] on both the x and y axis. Nonetheless, combining different affine transformations can lead to unrealistic cases. To avoid this, we simply look at the most extreme augmentation examples and consequently derive functional bounds for the transformation parameters. This explains why some of the chosen parameters may look peculiar. Chosen data augmentation parametrization for each class is available in Appendix \ref{appendix:data_aug} (Table \ref{tab:data_augmentation}).

Defining a proper data augmentation is expected to be crucial. One should notice that we design our strategies only once at the beginning: we do not optimize them with regards to the validation sets in order to improve performances. This is in order to simulate what a practitioner would do on a new use case. Note that the data augmentations are also designed as a compromise for all three methods. Indeed, a better data augmentation could probably be found for each individual method: for example, large translations on the object classes may improve Mahalanobis and KNN but could deteriorate PaDiM, since the latter does not discard the location information like the other two do via global average pooling.

\section{Experiments}
\label{sec:experiments}

\subsection{Dataset and metrics}
\label{subsec:datasets}

Since we are trying to solve a quality inspection task, we apply the methods to the MVTec AD dataset \cite{bergmann2019mvtec}, a now very common benchmark dataset in the anomaly detection field. This dataset is composed of 3629 training images and 1725 test images belonging to 15 different categories (or classes). The categories are often split into two types: textures (5 categories) and objects (10 categories). We train and validate our models on each category independently.

When reporting the area under the ROC curve (AUC), we solely focus on image-wise detection performances, and do not report any pixel-wise AUC metric, which usually aims to assess quality of the anomaly segmentation. This is because image-wise AUC is actually the main metric of interest in the industrial set-up. Additionally, pixel-wise AUC can easily reach high values, even when there still are false positives left on every single localization map, making the metric useless for choosing the right model to deploy to production.

\subsection{Implementation details}

For a fair comparison between KNN, Mahalanobis, and PaDiM, we use the same feature extractor for all three methods: we choose EfficientNet-B4 pre-trained on the ImageNet dataset\cite{DBLP:journals/corr/abs-1905-11946}, since it provided good results in previous anomaly detection works\cite{rippel2020modeling,defard2020padim}, as well as in our own network ablations (data not shown). For the feature levels, we decide to use the output of the blocks 4, 6, and 7 of the chosen feature extractor as our embedding layers, similarly to PaDiM original paper. These choices may be slightly detrimental to some methods and can explain the differences of performances with the ones reported in the original papers.

We implement all methods using Tensorflow / Keras for embedding the images with EfficientNetB4, and Scikit-Learn\cite{scikit-learn} for the traditional anomaly detection techniques (k nearest-neighors, covariance estimation, Mahalanobis distance). For data augmentation, we rely on Imgaug \cite{imgaug}. Contrarily to other previous approaches, we use almost no pre-processing, and only resize the images to the input resolution which was used by the feature extractor network for training. As we are using EfficientNet-B4, we resize all images to 380 x 380 pixels.

\subsection{How to measure robustness?}
\label{subsec:measuring}

In order to assess the methods robustness to low data regimes, we apply the following process. On a grid of sample sizes, we sample $N$ samples randomly without replacement. We fit a model with the chosen method on these $N$ samples. We replicate the process $M$ times for each of the chosen $N$ sample value in order to have a robust estimation of the performance for every $N$.

We iterate with models from all three methods, and use the same samples for each method to keep the comparison as fair as possible. We make the computation effective by encoding all training and evaluation embeddings with the feature extractor in advance. This way, for each new model fitting, only the later outlier detection module (KNN, Mahalanobis, or PaDiM) requires computation, which does not need to rely on a GPU or other intensive resources. In practice, for the MVTec dataset, we use a reproduction factor of $M =5$. We use sample sizes ranging from 5 to the maximum sample size, with a step of 5 for sample sizes $N$ below 50, and of 10 for sample sizes above. This is because increasing the sample size provides logarithmically diminishing returns. Over the MVTec dataset, we hence compute 3094 models for each method. The mere fact that we are able to fit and evaluate these numerous models is a strong argument in favor of the adoption of these methods, since they are blazing fast to train and do not require demanding compute resources.

In order to evaluate the robustness of the methods to the low data regime, we introduce a new metric that we call the “area under the AUC-percent curve”. The area under the AUC-percent curve is defined as the area under the curve that gives the AUC as a function of the percentage of the total original data that was used for training the corresponding model. Defining such a metric allows us to average it across all MVTec categories, since the latter do not share the same amount of total available training images.  

\subsection{Robustness to low data regimes}
\label{subsec:robustness}

Table \ref{tab:auc_max_sample_size} shows the AUC obtained for each method using all available training images of the MVTec dataset. For the Mahalanobis method, we compare the Empirical and the Ledoit estimators of the covariance matrix. Since our experiments show Ledoit perform betters in small setups, we used this estimator for PaDiM. Results per category are available in the Appendix \ref{appendix:detailed_results} (Table \ref{tab:complete_auc_max_sample_size}). The following can be noticed. First, the KNN approach is outperformed by all other methods. Second, the choice of the covariance matrix estimator in the Mahalanobis method has a rather low impact, with Ledoit only being slightly better than Empirical. This is expected since the amount of data is reasonably high, making the shrinking of the covariance matrix unnecessary. Finally, PaDiM appears clearly as the best method for textures, with an impressive AUC of 0.996. However, it is outperformed by Mahalanobis on object categories on average.

\begin{table}[h!]
    \caption{Averaged AUCs at maximum sample size on the MVTec AD dataset.}
    \label{tab:auc_max_sample_size}
    \begin{center}
    \begin{tabular}{l||c|c|c|c}
    & KNN & Mahalanobis & Mahalanobis & PaDiM \\
    & & (Empirical) & (Ledoit) & (Ledoit) \\
    \hline
    all         & 0.911 &      0.957 & \B 0.962 & 0.957 \\
    textures & 0.893 &      0.963 &   0.957 & \B 0.996 \\
    objects  & 0.919 &      0.954 &   \B 0.964 & 0.938 \\
    \end{tabular}
    \end{center}
\end{table}

\begin{figure}
    \centering
    \includegraphics[height=5cm]{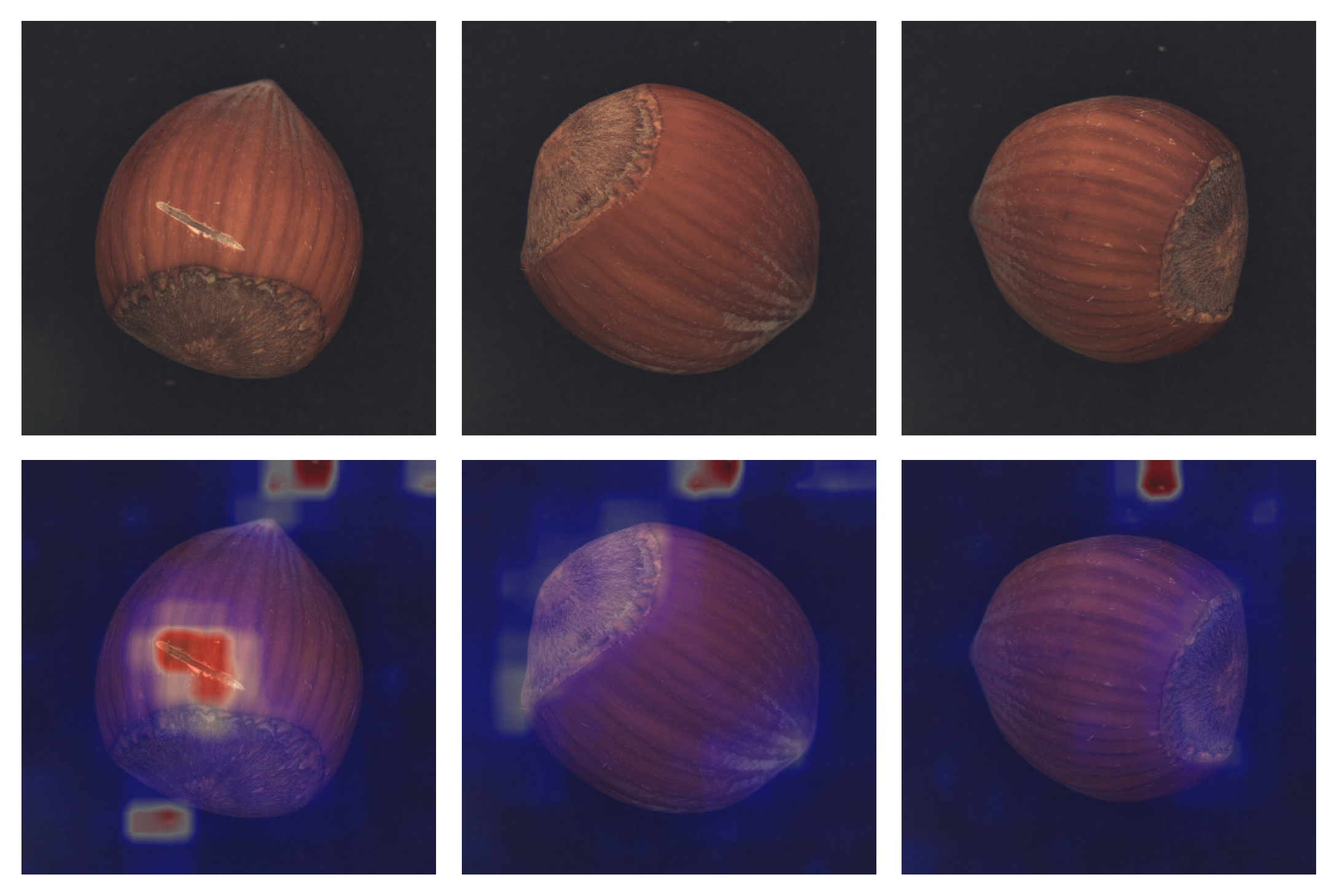}
    \caption{Images from the hazelnut category validation set (first row), with associated heatmaps obtained using PaDiM (second row). Notice how small local variations of the background induce false detections from PaDiM.}
    \label{fig:hazelnut}
\end{figure}

When looking at the per-class breakdown, the under-performance of PaDiM on objects is mostly due to the hazelnut category, which reaches an AUC of only 0.81. This is because the background is sometimes "polluted" by small particles for this category, as small coloured dots can be present on the images. Figure \ref{fig:hazelnut} gives an example of the issue. PaDiM tends to detect these as anomalies very strongly. Our hypothesis is that this is due to the fact that PaDiM models the normality in a local manner (“per patch”), while the other methods do not because of the global average pooling. Since the background is most of the time black, any slight local variation of the background will hence be detected as anomalous by PaDiM. Consequently, PaDiM is expected to be less robust to any kind of local perturbations compared to the other methods. Cropping, like it is done in the original PaDiM implementation\cite{defard2020padim}, can solve the issue. Indeed, when using a centered crop of size 600 x 600 pixels on the original images, it is possible to reach a much higher AUC of above 0.99. We hypothesize that data augmentation may also help to partially solve it, while methods making a more explicit use of spatial information such as DFR\cite{yang2020dfr} are expected to work even better in these kinds of pollution setups.

Note that these curves also partially answer the question "how much data do we need?". By looking at them, it is possible to know if the AUC can be improved by adding more data (ex: PaDiM on "capsule") or not (ex: PaDiM on "hazelnut", KNN on "metal nut", Mahalanobis on "pill", ...).

\begin{figure}
    \centering
    \includegraphics[height=5.5cm]{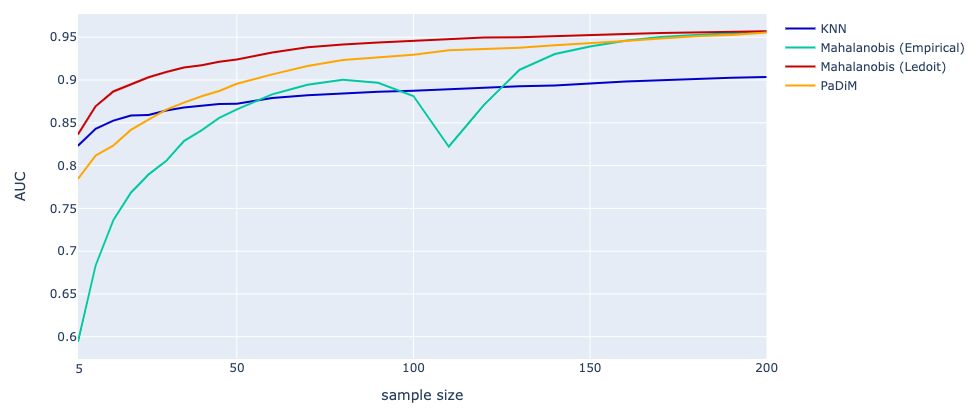}
    \caption{Average AUC across all MVTec classes, as a function of sample size. We removed the toothbrush category from the aggregation due to its low amount of datapoints.}
    \label{auc_per_sample_size}
\end{figure}

Table \ref{tab:agg_auc_perc} recapitulates the averaged areas under the AUC-percent curves for the MVTec AD dataset. The differences between approaches are clearer. Notably, these results demonstrate the interest of using the Ledoit covariance estimator instead of the Empirical one in the case of small training sample sizes. We complement these results with Figure \ref{auc_per_sample_size}, which displays the average AUC obtained on all MVTec classes as a function of the absolute sample size used (the breakdown for objects vs textures is available in figure \ref{auc_per_sample_size_textures} in appendix \ref{appendix:detailed_results}). Though Mahalanobis and PaDiM obtain similar performances when the training sample size is near 200 images, they behave very differently in low data regimes. In particular, Mahalanobis with Ledoit covariance estimator appears as the most robust to low data regimes of all methods, with an average AUC close to 0.85 when using only 5 training images: this is actually close to the performance of autoencoders with the use of the full training set \cite{bergmann2019mvtec}. In fact, with only 50 samples, the method reaches an average AUC of 0.927, beating most of the existing anomaly detection methods.

\begin{table}[h!]
    \caption{Averaged areas under the AUC-percent curve on the MVTec AD dataset.}
    \label{tab:agg_auc_perc}
    \begin{center}
    \begin{tabular}{l||c|c|c|c}
    & KNN & Mahalanobis & Mahalanobis & PaDiM \\
    & & (Empirical) & (Ledoit) & (Ledoit) \\
    \hline
    all         & 0.870 &      0.876 &   \B 0.920 &         0.899 \\
    textures & 0.864 &      0.882 &   0.925 &         \B 0.973 \\
    objects  & 0.873 &      0.873 &   \B 0.918 &         0.861 \\
    \end{tabular}
    \end{center}
\end{table}

As expected, using the Empirical covariance estimator is way more sensible to small training sample sizes. We also notice a consistent drop of performances when the sample size is in the neighborhood of the number of features of the first feature level (the output of the block4 level has dimension 112). A common interpretation for this phenomena is that the inverse of the empirical covariance estimator seems particularly poor at estimating the actual inverse of the true covariance matrix (a.k.a. the precision matrix) in the case where the number of samples is close to the number of features.\cite{7485996} For large covariance matrices with only few samples, the estimated matrix is singular, and the non-diagonal terms that actually represent correlation between features are indistinguishable from noise. While Ledoit estimator solves this by fixing the covariance matrix estimator, one could try to directly estimate the precision matrix in order to tackle the issue. Note that we expect the issue to arise for other feature levels if we were to use more data (since the other feature levels have a higher number of dimensions).

Comparatively, PaDiM seems to require more images to perform well. We hypothesize that this is due to the relative complexity of the method, as it estimates several covariance matrices instead of one. However, it is extremely stable for all textures: with only 5 images, the technique leads to an average AUC of 0.986, with an AUC of 1.0 on two texture categories: leather and carpet.

\subsection{Data augmentation for low data regimes}
\label{sec:data_augm_for_pre-trained}

\begin{figure}
    \centering
    \includegraphics[height=6cm]{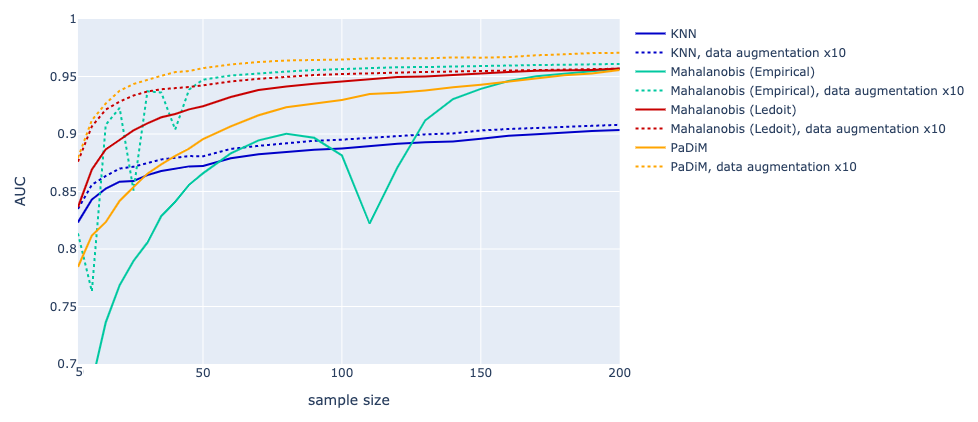}
    \caption{Average AUC across all MVTec classes as a function of the original sample size, for both the original data and its 10-times augmented counterpart. We removed the toothbrush category from the aggregation due to its low amount of datapoints.}
    \label{recap_aucs}
\end{figure}

We now show the results when augmenting the data by a factor of 10, making the effective sample size 11 times bigger than the original datasets. Figure \ref{recap_aucs} shows the AUC as a function of the original sample size for all methods, with and without the 10-times data augmentation. The areas under the AUC-percent curves are presented in Table \ref{tab:agg_auc_perc_with_da}. In general, all methods benefit from data augmentation. As hypothesized, because PaDiM is a more complex model, it generally benefits more from additional data ---and hence from data augmentation--- than the other methods. It can be noticed that using data augmentation makes the Ledoit covariance estimation slightly worse than the Empirical one as soon as the original sample size exceeds 50. Data augmentation almost always help in very low data regimes (less than 50 images). When dealing with a larger sample size (on average more than 200 images), data augmentation can however be detrimental, as shown by the Mahalanobis results on the grid, capsule or screw categories. Per class details can be found in Appendix \ref{appendix:detailed_results} (Figure \ref{frise_all}).

\begin{table}[h!]
    \caption{Averaged areas under the AUC-percent curve on the MVTec AD dataset, for both the original dataset and its 10-times augmented counterpart.}
    \label{tab:agg_auc_perc_with_da}
    \begin{center}
    \begin{tabular}{l||c|c|c|c|c|c|c|c}
     & \multicolumn{2}{c|}{KNN} & \multicolumn{2}{c|}{Mahalanobis} & \multicolumn{2}{c|}{Mahalanobis} & \multicolumn{2}{c}{PaDiM} \\
     & \multicolumn{2}{c|}{} & \multicolumn{2}{c|}{(Empirical)} & \multicolumn{2}{c|}{(Ledoit)} & \multicolumn{2}{c}{(Ledoit)} \\
    \hline
    augmentation factor &    0  &    10 &        0  &    10 &     0  &    10 &           0  &    10 \\
    \hline
    all         & 0.870 & 0.877 &     0.876 & 0.921 &  0.920 & 0.926 &        0.899 & \B 0.930 \\
    textures & 0.864 & 0.867 &     0.882 & 0.933 &  0.925 & 0.925 &        0.973 & \B  0.974 \\
    objects  & 0.873 & 0.882 &     0.873 & 0.916 &  0.918 & \B 0.927 &        0.861 & 0.908 \\
    \end{tabular}
    \end{center}
\end{table}

At maximum sample size (see also Table \ref{tab:auc_da_max} in Appendix \ref{appendix:detailed_results}), data augmentation does not help much Mahalanobis, as mentioned in the original paper \cite{rippel2020modeling}. However, we observe a significant improvement for PaDiM, which is mainly due to the gain on the hazelnut category (AUC increases from 0.81 to 0.89): this makes PaDiM with data augmentation the best performing method overall.

When focusing on small sample sizes, we can see that: a) using 30 images, data augmentation allows to gain 0.02 and 0.08 of AUC for Mahalanobis with Ledoit and PaDiM respectively, which corresponds to the performance obtained with 70 and 160 original images without data augmentation for the same methods respectively; and that b) using only 10 images and with data augmentation, we can reach an AUC higher than 0.9 for both Mahalanobis Ledoit and PaDiM, which is better than most existing autoencoder based approaches.\cite{bergmann2019mvtec} Corresponding tables are available in Appendix \ref{appendix:detailed_results} (Table \ref{tab:auc_da_30} and Table \ref{tab:auc_da_10}).

The results also validate our hypothesis from section \ref{subsec:robustness} that data augmentation can help with PaDiM failure cases such as hazelnut, for which better results are obtained with 10 images and a 10-factor data augmentation rather than with all of the available 391 images. To some extent, data augmentation, if chosen accordingly to semantic invariances, can translate into a location invariance to background noise, which, in the other methods, is achieved by global average pooling.

\subsection{How much data augmentation is needed?}

\begin{figure}
    \centering
    \includegraphics[height=4.8cm]{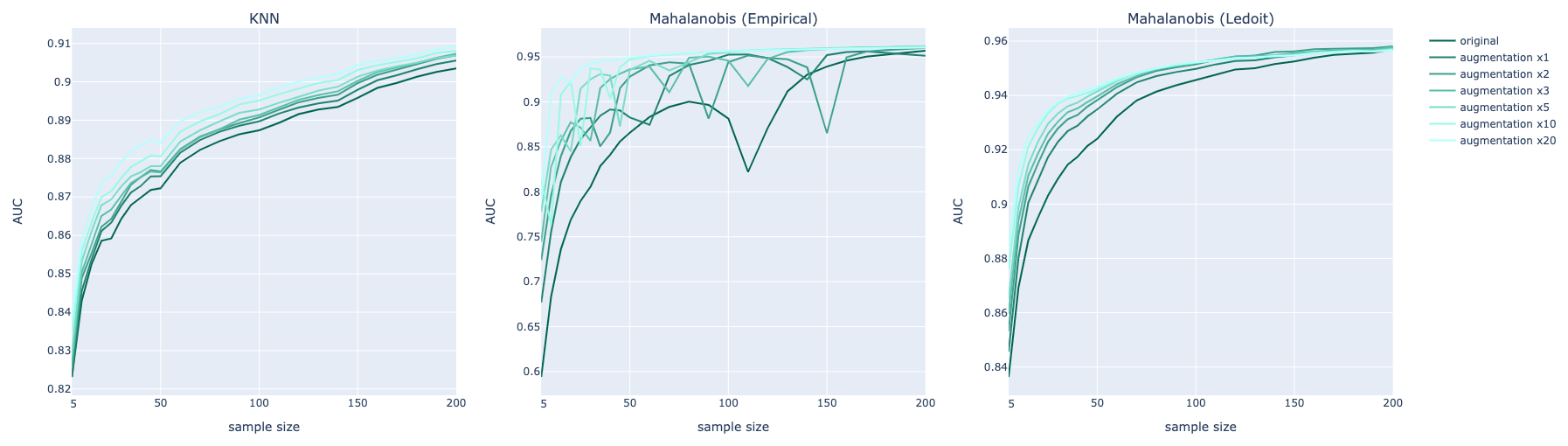}
    \caption{Average AUC across all MVTec classes as a function of original sample size, for the original data and for multiple factors of data augmentation (one curve per augmentation factor). We removed the toothbrush category from the aggregation due to its low amount of datapoints.}
    \label{ablation_aug_factor}
\end{figure}

On Figure \ref{ablation_aug_factor}, we show results for varying amounts of data augmentation applied: each curve represents the AUC as a function of the original sample size for a given augmentation factor. Because of compute time, we discard PaDiM from this experiment. As expected, increasing data augmentation provides diminishing returns. For Mahalanobis with Ledoit, it can even be harmful when dealing with more than 200 images. When working with more than 50 images, little difference is observed between a 10 and a 20 augmentation factor. Under 50 images, pushing the data augmentation further may still help improve the AUC slightly. For Mahalanobis with Empirical, a heavy data augmentation allows to avoid the performance drops discussed in \ref{subsec:robustness}. Finally, the KNN method always benefits from data augmentation, though its performance remains weaker than that of the other methods. This may be because the KNN method is by construction more dependent on the number of training datapoints than the other methods, in the hypothesis that a larger number of datapoints is required to properly separate the normal distribution from the anomalies (no distribution is estimated in the KNN case).

\begin{figure}
    \centering
    \includegraphics[height=6cm]{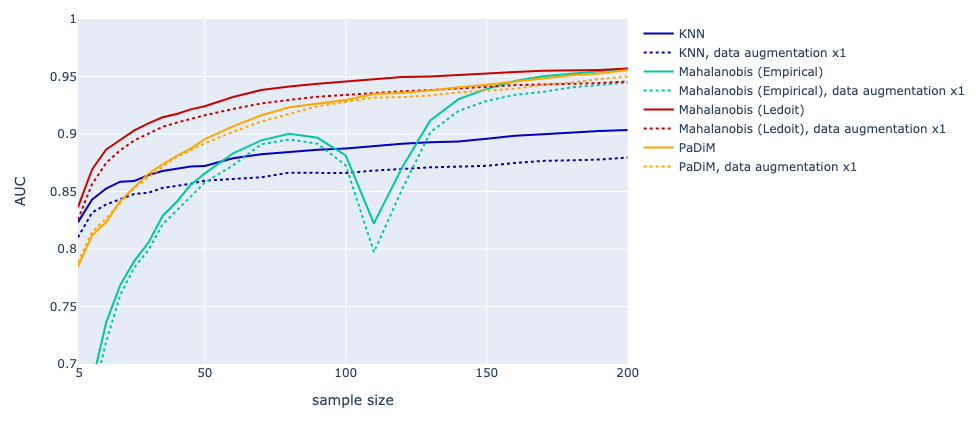}
    \caption{Average AUC across all MVTec classes as a function of the original sample size, for both the original data and its 1-time augmented counterpart (removing the original datapoints). We removed the toothbrush category from the aggregation due to its low amount of datapoints.}
    \label{fig:x1_data_augm}
\end{figure}

One other question that arises is the relationship between data augmentation and a possible increase in training time. For the KNN and the Mahalanobis techniques, training time is mostly spent on image embedding, which grows linearly with the amount of training images. To minimize training time, one could augment each image once and discard the original images for training (thus only keeping the augmented ones). That way, training time would remain stable while diversity of the training set would increase. Figure \ref{fig:x1_data_augm} shows that this strategy does not work in practice. We hypothesize that this is due to the shift induced by our data augmentation.

\section{Conclusion}
\label{sec:conclusion}

In this work, we compared three different approaches that all make use of deep pre-trained feature extractors. We showed that although these methods are in general quite robust to small sample sizes, they can still be improved further by using data augmentation. Our results show that without training any deep learning model, one can achieve very good results, even with few images. Notably, with only 10 images per class and in less than 2 seconds of training, we achieve better results than most auto encoders based techniques on the MVtec dataset.

A question that arises for real time applications like automated quality control is the trade-off between the increase in training time caused by the data augmentation, and the possibility to manually gather and annotate more images. We argue that the latter cannot be done for very small series of parts, since the whole production in these cases may not exceed 500 parts, most of which a human operator can inspect instead of annotate.

One of the downsides of our current methodology is that the data augmentation policy has to be designed manually. A possibility in the future could be to use reinforcement learning to generate optimal augmentation policies: this could be made possible thanks to the low training time that the pre-trained networks based methods offer. However, as we have shown, the data augmentation policy for a given sample size may not be optimal for another one, which increases the complexity of the search policy.

\bibliography{report} 
\bibliographystyle{spiebib} 

\pagebreak
\appendix
\section{Data Augmentation details}
\label{appendix:data_aug}
\begin{table}[h!]
    \caption{Data augmentation details per MVTec AD category. “Rotate (90°s)” indicates rotations of multiples of 90° (hence 90°, 180°, and 270°). Each data augmentation was manually and visually designed by a human.}
    \label{tab:data_augmentation}    
    \begin{center}
    \begin{tabular}{l|c|c|c|c|c|c|c|c}
                  & Flip & Flip & Translate  & Rotate & Rotate & Zoom & Add &   Multiply \\
               
                & (horizontal) & (vertical) & ($\pm x, \pm y$) & (range in °) & (90°s) & (range) & (range) & (range) \\
    \hline
              carpet &               Yes &            Yes &                (5, 5) &             - &         Yes &            - &  (-10, 10) &  (0.9, 1.1) \\
                tile &               Yes &            Yes &                (5, 5) &             - &         Yes &            - &  (-10, 10) &  (0.9, 1.1) \\
             leather &               Yes &            Yes &                (5, 5) &             - &         Yes &            - &  (-10, 10) &  (0.9, 1.1) \\
                grid &               Yes &            Yes &                (5, 5) &             - &         Yes &            - &  (-10, 10) &  (0.9, 1.1) \\
                wood &               Yes &            Yes &                (5, 5) &       (-2, +2) &         - &            - &  (-10, 10) &  (0.9, 1.1) \\
             capsule &               - &            - &              (10, 10) &       (-3, +3) &         - &  (0.98, 1.02) &  (-10, 10) &  (0.9, 1.1) \\
               cable &               - &            - &              (10, 10) &       (-5, +5) &         - &           &  (-10, 10) &  (0.9, 1.1) \\
                pill &               - &            - &              (10, 10) &       (-3, +3) &         - &  (0.98, 1.02) &  (-10, 10) &  (0.9, 1.1) \\
          transistor &               Yes &            - &                (5, 5) &       (-2, +2) &         - &            - &  (-10, 10) &  (0.9, 1.1) \\
           metal nut &               - &            - &              (10, 10) &     (-10, +10) &         Yes &            - &  (-10, 10) &  (0.9, 1.1) \\
          toothbrush &               Yes &            - &              (10, 10) &             - &         - &            - &  (-10, 10) &  (0.9, 1.1) \\
               screw &               Yes &            Yes &              (10, 10) &     (-10, +10) &         Yes &  (0.98, 1.02) &  (-10, 10) &  (0.9, 1.1) \\
            hazelnut &               Yes &            Yes &              (10, 10) &     (-20, +20) &         Yes &  (0.98, 1.02) &  (-10, 10) &  (0.9, 1.1) \\
              zipper &               Yes &            - &               (30, 0) &             - &         - &            - &  (-10, 10) &  (0.9, 1.1) \\
              bottle &               - &            - &                (5, 5) &     (-10, +10) &         Yes &            - &  (-10, 10) &  (0.9, 1.1) \\
    \end{tabular}
    \end{center}
\end{table}
\pagebreak

\section{Results details}
\label{appendix:detailed_results}
\begin{table}[h!]
    \caption{AUC at maximum sample size per MVTec AD category.}
    \label{tab:complete_auc_max_sample_size}
    \begin{center}
    \begin{tabular}{l||c|c|c|c}
     &   KNN &  Mahalanobis &  Mahalanobis &  PaDiM \\
     &    &  (Empirical) &  (Ledoit) & (Ledoit) \\
    \hline
    carpet     & 0.982 &      \B 1.000 &   \B 1.000 &         \B 1.000 \\
    tile       & 0.995 &      \B 1.000 &   \B 1.000 &         0.997 \\
    leather    & 0.994 &      \B 1.000 &   \B 1.000 &         \B 1.000 \\
    grid       & 0.625 &      0.833 &   0.825 &         \B 0.988 \\
    wood       & 0.870 &      \B 0.982 &   0.961 &         0.993 \\
    capsule    & 0.948 &      0.960 &   \B 0.966 &         0.959 \\
    cable      & 0.938 &      0.975 &   0.977 &         \B 0.980 \\
    pill       & 0.816 &      0.811 &   0.893 &         \B 0.947 \\
    transistor & 0.927 &      \B 0.976 &   \B 0.976 &         0.949 \\
    metal nut  & 0.868 &      0.952 &   \B 0.956 &         0.953 \\
    toothbrush & 0.919 &      \B 0.983 &   0.961 &         0.864 \\
    screw      & 0.884 &      0.911 &   0.937 &         \B 0.964 \\
    hazelnut   & 0.952 &      \B 0.997 &   \B 0.997 &         0.809 \\
    zipper     & 0.945 &      \B 0.974 &   \B 0.974 &         0.956 \\
    bottle     & 0.996 &      \B 1.000 &   \B 1.000 &         0.994 \\
    \end{tabular}
    \end{center}
\end{table}

\begin{table}[h!]
    \caption{Area under the AUC-percent curve per MVTec AD category.}
    \label{tab:auc_percent}
    \begin{center}
    \begin{tabular}{l||c|c|c|c}
     &   KNN &  Mahalanobis &  Mahalanobis &  PaDiM \\
     &    &  (Empirical) &  (Ledoit) & (Ledoit) \\
    \hline
    carpet     & 0.964 &      0.925 &   0.980 &         \B 0.982 \\
    tile       & 0.969 &      0.961 &   \B 0.976 &         0.968 \\
    leather    & 0.972 &      0.960 &   0.979 &         \B 0.980 \\
    grid       & 0.572 &      0.710 &   0.762 &         \B 0.967 \\
    wood       & 0.843 &      0.857 &   0.928 &         \B 0.970 \\
    capsule    & 0.892 &      0.886 &   \B 0.926 &         0.849 \\
    cable      & 0.914 &      0.867 &   \B 0.934 &         0.929 \\
    pill       & 0.778 &      0.809 &   0.850 &         \B 0.878 \\
    transistor & 0.891 &      0.876 &   \B 0.931 &         0.880 \\
    metal nut  & 0.843 &      0.869 &   \B 0.912 &         0.855 \\
    toothbrush & 0.834 &      0.834 &   \B 0.875 &         0.762 \\
    screw      & 0.773 &      0.761 &   0.852 &         \B 0.855 \\
    hazelnut   & 0.919 &      0.938 &   \B 0.976 &         0.721 \\
    zipper     & 0.918 &      0.928 &   \B 0.944 &         0.922 \\
    bottle     & 0.967 &      0.965 &   \B 0.975 &         0.964 \\
    \end{tabular}
    \end{center}
\end{table}

\begin{figure}
    \centering
    \includegraphics[height=7.5cm]{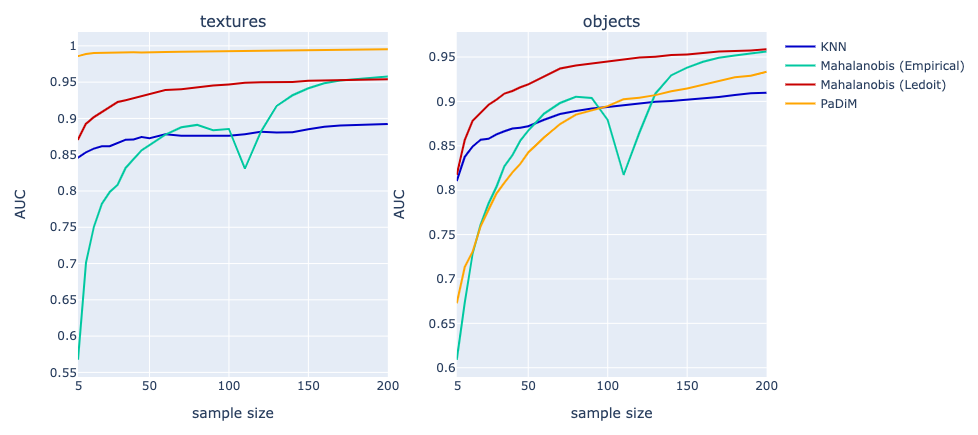}
    \caption{Average AUC across all MVTec AD texture and object categories as a function of sample size. We removed the toothbrush category from the aggregation due to its low amount of datapoints.}
    \label{auc_per_sample_size_textures}
\end{figure}

\begin{figure}
    \centering
    \includegraphics[height=21cm]{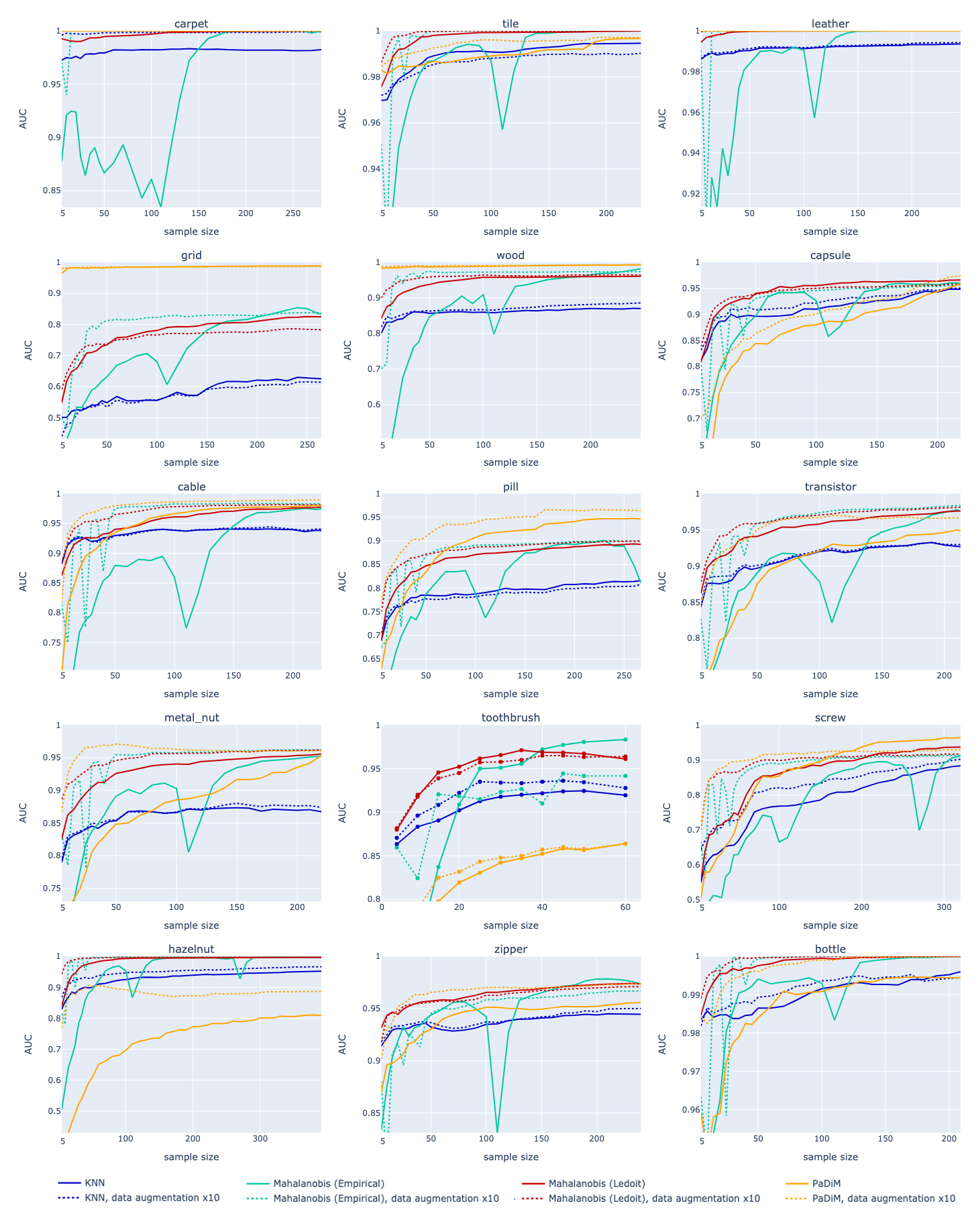}
    \caption{AUC as a function of the original sample size for all MVTec categories.}
    \label{frise_all}
\end{figure}

\begin{table}[h!]
\caption{Averaged AUCs at maximum sample size for MVTec, for both the original data and its 10-times augmented counterpart.}
\label{tab:auc_da_max}
\begin{center}
    \begin{tabular}{l||c|c|c|c|c|c|c|c}
     & \multicolumn{2}{c|}{KNN} & \multicolumn{2}{c|}{Mahalanobis} & \multicolumn{2}{c|}{Mahalanobis} & \multicolumn{2}{c}{PaDiM} \\
     & \multicolumn{2}{c|}{} & \multicolumn{2}{c|}{(Empirical)} & \multicolumn{2}{c|}{(Ledoit)} & \multicolumn{2}{c}{(Ledoit)} \\
    \hline
    augmentation factor &    0  &    10 &        0  &    10 &     0  &    10 &           0  &    10 \\
    \hline
    all         & 0.911 & 0.915 &     0.957 & 0.961 &  0.962 & 0.959 &        0.957 & \B 0.966 \\
    textures & 0.893 & 0.897 &     0.963 & 0.962 &  0.957 & 0.949 &        \B 0.996 & 0.995 \\
    objects  & 0.919 & 0.924 &     0.954 & 0.961 &  \B 0.964 & 0.963 &        0.938 & 0.951 \\
    \end{tabular}
\end{center}
\end{table}

\begin{table}
\caption{Averaged AUCs at original sample size 30 for MVTec, for both the original data and its 10-times augmented counterpart.}
\label{tab:auc_da_30}
\begin{center}
    \begin{tabular}{l||c|c|c|c|c|c|c|c}
     & \multicolumn{2}{c|}{KNN} & \multicolumn{2}{c|}{Mahalanobis} & \multicolumn{2}{c|}{Mahalanobis} & \multicolumn{2}{c}{PaDiM} \\
     & \multicolumn{2}{c|}{} & \multicolumn{2}{c|}{(Empirical)} & \multicolumn{2}{c|}{(Ledoit)} & \multicolumn{2}{c}{(Ledoit)} \\
    \hline
    augmentation factor &    0  &    10 &        0  &    10 &     0  &    10 &           0  &    10 \\
    \hline
    all         & 0.868 & 0.879 &     0.815 & 0.936 &  0.913 & 0.938 &        0.864 & \B 0.941 \\
    textures & 0.867 & 0.872 &     0.809 & 0.948 &  0.923 & 0.936 &        0.990 & \B 0.993 \\
    objects  & 0.868 & 0.882 &     0.818 & 0.930 &  0.908 & \B 0.939 &        0.801 & 0.915 \\
    \end{tabular}
\end{center}
\end{table}

\begin{table}
\caption{Averaged AUCs at original sample size 10 for MVTec, for both the original data and its 10-times augmented counterpart.}
\label{tab:auc_da_10}
\begin{center}
    \begin{tabular}{l||c|c|c|c|c|c|c|c}
     & \multicolumn{2}{c|}{KNN} & \multicolumn{2}{c|}{Mahalanobis} & \multicolumn{2}{c|}{Mahalanobis} & \multicolumn{2}{c}{PaDiM} \\
     & \multicolumn{2}{c|}{} & \multicolumn{2}{c|}{(Empirical)} & \multicolumn{2}{c|}{(Ledoit)} & \multicolumn{2}{c}{(Ledoit)} \\
    \hline
    augmentation factor &    0  &    10 &        0  &    10 &     0  &    10 &           0  &    10 \\
    \hline
    all         & 0.846 & 0.859 &     0.687 & 0.767 &  0.872 & \B 0.908 &        0.809 & 0.904 \\
    textures & 0.853 & 0.857 &     0.701 & 0.788 &  0.892 & 0.912 &        0.989 & \B  0.991 \\
    objects  & 0.842 & 0.859 &     0.680 & 0.757 &  0.863 & \B 0.905 &        0.719 & 0.860 \\
    \end{tabular}
\end{center}
\end{table}

\end{document}